\documentclass[runningheads]{llncs}
\usepackage[T1]{fontenc}
\usepackage{graphicx}
\usepackage{amsmath}
\usepackage{multirow}
\usepackage{amssymb}
\usepackage[colorlinks=true, linkcolor=black, urlcolor=blue]{hyperref}
\usepackage{orcidlink}
\usepackage{algorithm}
\usepackage{algpseudocode}
\usepackage{todonotes}
 \usepackage{booktabs}
\begin{document}
\title{HierAdaptMR: Cross-Center Cardiac MRI Reconstruction with Hierarchical Feature Adapters}

\author{Ruru Xu, Ilkay Oksuz\orcidlink{0000-0001-6478-0534}}
\authorrunning{Ruru Xu et al.}
\institute{Computer Engineering Department, Istanbul Technical University, Istanbul, Turkey
\email{xu21@itu.edu.tr}\\
\url{https://github.com/Ruru-Xu/HierAdaptMR}
}
 
\maketitle            
\begin{abstract}
Deep learning-based cardiac MRI reconstruction faces significant domain shift challenges when deployed across multiple clinical centers with heterogeneous scanner configurations and imaging protocols. We propose HierAdaptMR, a hierarchical feature adaptation framework that addresses multi-level domain variations through parameter-efficient adapters. Our method employs Protocol-Level Adapters for sequence-specific characteristics and Center-Level Adapters for scanner-dependent variations, built upon a variational unrolling backbone. A Universal Adapter enables generalization to entirely unseen centers through stochastic training that learns center-invariant adaptations. The framework utilizes multi-scale SSIM loss with frequency domain enhancement and contrast-adaptive weighting for robust optimization. Comprehensive evaluation on the CMRxRecon2025 dataset spanning 5+ centers, 10+ scanners, and 9 modalities demonstrates superior cross-center generalization while maintaining reconstruction quality.

\end{abstract}

\section{Introduction}

Accelerated cardiac MRI reconstruction using deep learning has achieved remarkable single-center performance, yet deployment across multiple clinical sites reveals significant domain shift challenges\cite{wang2024cmrxrecon}. The CMRxRecon2025\cite{wang2025towards}\cite{wang2025cmrxrecon2024} challenge exposes this limitation by evaluating models on entirely unseen centers, reflecting real-world clinical scenarios where reconstruction systems must generalize across hierarchical domain variations. These variations span multiple levels: scanner-level differences including field strengths from 1.5T to 5.0T, vendor implementations from GE, Philips, Siemens, and UIH, along with diverse coil configurations; protocol-level variations encompassing cine, LGE, T1/T2 mapping, and perfusion sequences with distinct reconstruction requirements\cite{wang2019black}\cite{lyu2024stadnet}; and acquisition-level differences featuring uniform 8× to 24× acceleration with k-t Gaussian and radial sampling patterns across various anatomical orientations.

Existing domain adaptation approaches either require full model retraining for each target site or apply post-processing harmonization that may remove diagnostic information. Parameter-efficient adaptation techniques, successful in computer vision, remain unexplored for medical image reconstruction where both reconstruction quality and computational efficiency are critical. Traditional MRI reconstruction methods have evolved from classical approaches\cite{qin2018convolutional}\cite{qin2021complementary} to advanced deep learning frameworks incorporating GANs\cite{lyu2020parallel}, and federated learning\cite{lyu2023adaptive}.
PromptMR\cite{xin2023fill} introduced a pioneering framework for dynamic and multi-contrast MRI reconstruction through k-space filling and image refinement with learnable prompts. The subsequent PromptMR-plus\cite{xin2024rethinking} further enhanced this approach by rethinking deep unrolled models for accelerated MRI reconstruction. These advances have inspired numerous applications in cardiac MRI, including recent developments in the CMRxRecon2024 Challenge where PromptMR-based methods demonstrated competitive performance across various reconstruction tasks\cite{xu2024hypercmr}\cite{anvari2024all}\cite{lyu2024upcmr}. Additional works have extended PromptMR's capabilities to reinforcement learning-based sampling optimization\cite{xu2025reinforcement}, progressive divide-and-conquer strategies\cite{wang2024progressive}, and quality assessment\cite{nabavi2024statistical}.

Building upon the previous work HyperCMR\cite{xu2024hypercmr}, which achieved 5th place in the CMRxRecon2024 Challenge, we propose a hierarchical feature adaptation framework for cross-center cardiac MRI reconstruction. Our approach addresses the fundamental challenge of deploying reconstruction models across diverse clinical environments while maintaining reconstruction quality across different centers. The framework extends HyperCMR\cite{xu2024hypercmr} with lightweight adapters that perform feature-level adaptation, enabling end-to-end optimization of the entire network architecture. The architecture addresses scanner variations through center-specific networks and protocol variations through contrast-specific networks. A hybrid training strategy combines specialized and universal adaptation, enabling generalization to unseen centers while maintaining performance on known domains.

Our contributions include: 
\begin{itemize}
    \item A hierarchical feature adaptation framework addressing both scanner-level and protocol-level domain shifts through parameter-efficient adapters while preserving pre-trained model capabilities.
    \item A novel hybrid training strategy with probability-based adapter selection that enables effective generalization to unseen centers through universal adapter learning.
    \item Comprehensive evaluation on the CMRxRecon2025 dataset spanning 5+ centers, 10+ scanners, and 9 imaging modalities, demonstrating superior cross-center generalization.
\end{itemize}

\section{Method}

\subsection{Network Architecture}

\begin{figure}[t]
    \centering
    \includegraphics[width=1\linewidth]{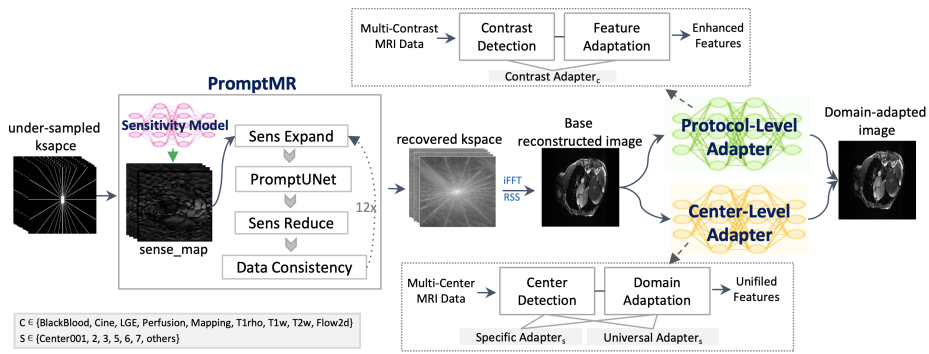}
    \caption{Hierarchical Feature Adaptation Framework. The architecture integrates HyperCMR's variational unrolling with dual-level adaptation: Protocol-Level Adapters address sequence-dependent variations while Center-Level Adapters handle scanner-specific differences. A Universal Adapter enables generalization to entirely unseen centers during evaluation.}
    \label{fig:fig-method}
\end{figure}

As shown in Figure \ref{fig:fig-method}, we propose a hierarchical feature adaptation framework for cross-center cardiac MRI reconstruction that addresses multi-level domain shifts encountered when deploying models to entirely unseen clinical centers, as required by the CMRxRecon2025 challenge.

\textbf{Reconstruction Backbone:} Comprehensive optimization of HyperCMR and PromptMR-plus. HyperCMR\cite{xu2024hypercmr}, which implements variational network unrolling with learnable prompts. The iterative k-space refinement follows:

$$\mathbf{k}^{(t+1)} = \mathbf{k}^{(t)} - \lambda_t \odot \mathcal{A}^H (\mathcal{M} (\mathcal{A} \mathbf{k}^{(t)} - \mathbf{y}) + \mathcal{R}_{\theta_t}(\mathcal{A} \mathbf{k}^{(t)}))$$

where $\mathcal{A}$ denotes the forward encoding operator, $\mathcal{M}$ represents the sampling mask, $\mathbf{y}$ is the acquired k-space data, and $\mathcal{R}_{\theta_t}$ is the learnable regularization term at unrolling step $t$. 

\begin{figure}[h]
\centering
\includegraphics[width=1\linewidth]{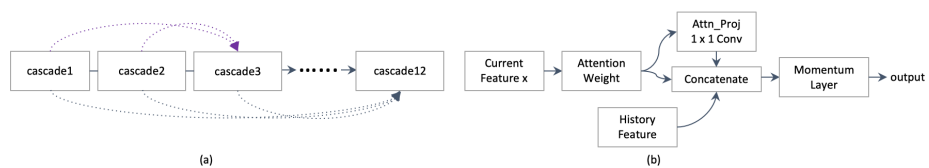}
\caption{Historical feature fusion architecture for enhanced cascade-based reconstruction. The proposed framework addresses feature information loss in traditional cascade networks through adaptive attention mechanisms and multi-scale memory integration.}
\label{fig:fig-history}
\end{figure}

We combine HyperCMR and PromptMR-plus \cite{xin2024rethinking} as our backbone. But the PromptMR-plus identified a key limitation: cascade networks discard valuable intermediate features between stages. Figure \ref{fig:fig-history} shows our solution.
We introduce a historical feature memory that preserves and selectively reuses accumulated representations across cascade stages. The framework employs adaptive attention to dynamically weight current against historical features, maintains a multi-scale memory bank preserving intermediate outputs, and uses momentum-based integration for stable convergence.
For cascade stage $i$, enhanced features are computed as:

$$\mathbf{F}_i^{\text{enhanced}} = \alpha_i \odot \mathbf{F}_i^{\text{current}} + (1-\alpha_i) \odot \mathcal{M}(\{\mathbf{F}_j\}_{j=1}^{i-1})$$

where $\alpha_i$ represents learned attention weights, $\mathbf{F}_i^{\text{current}}$ denotes current stage features, and $\mathcal{M}(\cdot)$ is the memory aggregation function.

This transforms traditional feed-forward cascades into recurrent-like structures that maintain long-term dependencies while preserving computational efficiency, yielding improved reconstruction fidelity for fine anatomical structures.

\textbf{Hierarchical Adaptation Strategy.} To address the challenge of generalizing to entirely unseen centers with unknown scanner configurations and protocols, our adaptation operates on reconstructed images $\mathbf{I}_{\text{base}} = \text{RSS}(|\mathcal{F}^{-1}(\mathbf{k}_{\text{recon}})|)$ through two complementary levels:

\textit{Protocol-Level Adaptation} employs specialized networks $\mathcal{T}_p^{(\text{protocol})}$ for different imaging sequences (cine, LGE, T1/T2 mapping, perfusion) to adapt to sequence-specific characteristics, including temporal dynamics, contrast behavior, and reconstruction requirements.

\textit{Center-Level Adaptation} uses dedicated networks $\mathcal{T}_c^{(\text{center})}$ for known training centers to capture scanner-dependent variations such as field strengths (1.5T-5.0T), vendor differences (GE, Philips, Siemens, UIH), coil configurations, and acquisition patterns.

\textbf{Adapter Architecture.} Each adapter implements a lightweight residual learning structure:

$$\mathcal{T}(\mathbf{x}) = \mathbf{x} + \alpha \cdot \tanh(\mathbf{W}_3 \ast \phi(\mathbf{W}_2 \ast \phi(\mathbf{W}_1 \ast \mathcal{N}(\mathbf{x}))))$$

where $\phi(\cdot)$ combines batch normalization and ReLU activation, $\mathcal{N}(\cdot)$ applies instance normalization for input standardization. The architecture uses three convolutional layers with channel dimensions [C, C/4, C/16, 1], where C is the input channel dimension, maintaining spatial dimensions while progressively reducing feature complexity. The scaling factor $\alpha$ is initialized to small values (e.g., 0.1) to ensure stable adaptation without disrupting the backbone's reconstruction capability.

\textbf{Universal Adapter for Unseen Center Generalization.} The key challenge of CMRxRecon2025 is to generalize to the entirely unseen centers during evaluation. To address this, we introduce a Universal Adapter $\mathcal{T}_u$ that learns center-invariant adaptations. During training on known centers, we employ stochastic adapter selection for center-level adaptation:

$$\mathcal{T}_{\text{center-active}} = \begin{cases}
\mathcal{T}_c^{(\text{center})} & \text{with probability } p \\
\mathcal{T}_u & \text{with probability } 1-p
\end{cases}$$

This mixed training strategy forces $\mathcal{T}_u$ to learn generalizable adaptations by experiencing diverse scanner and protocol variations from multiple known centers.

\textbf{Adapter Selection Strategy.} During training, the active adapter combination is determined as:
$$\mathcal{T}_{\text{active}} = \mathcal{T}_p^{(\text{seq})} \circ \mathcal{T}_{\text{center-active}}$$
where $\mathcal{T}_p^{(\text{seq})}$ is deterministically selected based on the imaging sequence, and $\mathcal{T}_{\text{center-active}}$ follows the stochastic selection described above. During evaluation on unseen centers, $\mathcal{T}_{\text{center-active}} = \mathcal{T}_u$, while protocol-specific adapters are selected based on the known imaging sequence.

The final adapted reconstruction is obtained as: $\mathbf{I}_{\text{final}} = \mathcal{T}_{\text{active}}(\mathbf{I}_{\text{base}})$.

\textbf{End-to-End Optimization.} The entire framework is trained end-to-end to optimize both reconstruction quality and cross-center generalization:

$$\min_{\theta_{\text{base}}, \Theta_{\text{adapt}}} \mathbb{E}_{(\mathbf{x},\mathbf{y}), \mathcal{T}_{\text{active}}} \left[ \ell(\mathcal{T}_{\text{active}}(\mathcal{F}(\mathbf{x}; \theta_{\text{base}})), \mathbf{y}) + \beta \sum_{i} \|\Theta_{\text{adapt}}^{(i)}\|_F^2 \right]$$

where $\ell(\cdot,\cdot)$ combines reconstruction loss terms (e.g., L1 and SSIM losses), $\mathcal{F}(\mathbf{x}; \theta_{\text{base}})$ represents the HyperCMR backbone, $\theta_{\text{base}}$ denotes backbone parameters, and $\Theta_{\text{adapt}} = \{\mathcal{T}_c^{(\text{center})}, \mathcal{T}_p^{(\text{protocol})}, \mathcal{T}_u\}$ represents all adapter parameters. The expectation over $\mathcal{T}_{\text{active}}$ accounts for the stochastic adapter selection during training.

\textbf{Adapter Functionality.} It is important to note that the proposed adapters serve a specific purpose distinct from traditional image enhancement techniques. The Protocol-Level Adapters process initial reconstructed images to address sequence-specific characteristics across different imaging modalities (cine, LGE, T1/T2 mapping, perfusion), while Center-Level Adapters handle scanner-dependent variations including field strength differences, vendor-specific implementations, and coil configuration disparities. These adapters are designed for domain adaptation rather than image enhancement tasks such as denoising or super-resolution. The primary resolution enhancement occurs within the HyperCMR backbone through the reconstruction process from undersampled k-space to fully reconstructed images. The adapters operate as post-processing modules that perform feature-level domain alignment to ensure consistent reconstruction quality across diverse clinical environments, thereby addressing the fundamental challenge of cross-center generalization in cardiac MRI reconstruction.

\subsection{Loss Function} 
Our loss function design addresses the specific challenges of cardiac MRI reconstruction across diverse imaging contrasts and acquisition protocols. Rather than relying solely on pixel-wise metrics, we develop a comprehensive SSIM-based framework that operates across multiple spatial and frequency domains to ensure robust reconstruction quality in heterogeneous clinical environments.

\textbf{Multi-Scale SSIM Framework.} We leverage multi-scale structural similarity assessment to capture anatomical features at different resolution levels:

$$\mathcal{L}_{\text{MS-SSIM}} = \frac{1}{|\mathcal{S}|} \sum_{s \in \mathcal{S}} \frac{1}{s} \left(1 - \text{SSIM}(\mathcal{D}_s(\hat{\mathbf{I}}), \mathcal{D}_s(\mathbf{I}))\right)$$

where $\mathcal{S}$ represents downsampling scales, $\mathcal{D}_s(\cdot)$ denotes average pooling with kernel size $s$, and the scale-dependent weighting $1/s$ emphasizes fine-scale details while preserving coarse structural features. This multi-resolution approach maintains both global anatomical coherence and local textural fidelity across different imaging protocols.

\textbf{Frequency Domain Enhancement.} To complement spatial evaluation, we incorporate frequency domain analysis that captures spectral characteristics essential for medical image quality:

$$\mathcal{L}_{\text{Freq-SSIM}} = 1 - \text{SSIM}(\mathcal{N}(|\mathcal{F}(\hat{\mathbf{I}})|), \mathcal{N}(|\mathcal{F}(\mathbf{I})|))$$

where $\mathcal{F}(\cdot)$ represents the 2D Fourier transform and $\mathcal{N}(\cdot)$ applies normalization. This component ensures preservation of both low-frequency anatomical structures and high-frequency textural details critical for clinical interpretation.

\textbf{Contrast-Adaptive Weighting.} Recognizing the heterogeneous nature of cardiac MRI protocols, we implement contrast-adaptive weighting that adjusts loss contributions based on imaging sequence characteristics:

$$\mathcal{L}_{\text{total}} = w_{\text{base}}^c \cdot \mathcal{L}_{\text{SSIM}} + w_{\text{ms}}^c \cdot \mathcal{L}_{\text{MS-SSIM}} + w_{\text{freq}}^c \cdot \mathcal{L}_{\text{Freq-SSIM}}$$

The weights $\{w_{\text{base}}^c, w_{\text{ms}}^c, w_{\text{freq}}^c\}$ are optimized for each protocol: cine sequences emphasize base SSIM for temporal consistency, LGE sequences employ elevated multi-scale weighting for contrast preservation, and mapping sequences prioritize spatial fidelity with reduced frequency contribution. This adaptive strategy ensures appropriate response to the specific requirements of different cardiac imaging modalities while maintaining computational efficiency for clinical deployment.

\section{Experiments and Results Analysis}
\subsection{Experiments}

\noindent\textbf{Dataset and Experimental Setup:} We evaluate on the CMRxRecon2025 dataset comprising 5+ medical centers with 10+ scanners from major vendors (GE, Philips, Siemens, UIH) across field strengths of 1.5T-5.0T. The dataset includes 9 modalities (Cine, LGE, Mapping, Perfusion, T1rho, Flow2d, BlackBlood, T1w, T2w) with multiple anatomical views. Raw k-space data is converted from MATLAB to HDF5 format with RSS ground truth reconstruction. We implement a rigorous center-based splitting strategy where one complete center is intelligently selected as validation set to simulate unseen domain evaluation. The validation center is automatically chosen to achieve approximately 15\% data split while ensuring zero patient overlap between training and validation sets. This splitting methodology guarantees that the model never sees any data from the validation center during training, providing a realistic assessment of cross-center generalization capability.

\noindent\textbf{Implementation Details:} Training employs PyTorch on NVIDIA A100 80GB with mixed-precision and gradient accumulation (8 steps). Our framework utilizes the HyperCMR backbone (12 cascades, $48 \rightarrow [72,96,120]$ features) with lightweight adaptation modules in an end-to-end training paradigm. We employ differentiated learning rates for optimal convergence: backbone parameters at $2 \times 10^{-4}$, adapter modules at $4 \times 10^{-4}$ using AdamW optimizer (weight decay=$1 \times 10^{-4}$). The higher learning rate for adapters enables faster domain-specific feature learning while maintaining the stability of the reconstruction backbone. Training uses cosine annealing scheduling ($\eta_{\text{min}}=1 \times 10^{-6}$) with early stopping (patience=2 epochs) based on validation SSIM. The framework processes 5 adjacent temporal slices with circular padding and employs hierarchical stratified sampling (30\% subset) maintaining Center→Vendor→Modality→Patient diversity structure.

\subsection{Results Analysis}
The comprehensive evaluation results warrant detailed examination across both quantitative metrics and qualitative assessments. We analyze these findings through multiple perspectives to demonstrate the robustness and clinical applicability of our approach.

\begin{table}[htbp]
\centering
\caption{Quantitative cross-center performance evaluation on CMRxRecon2025 dataset. Results show SSIM and PSNR metrics comparing HyperCMR baseline with our HierAdaptMR method across different center configurations and scanner vendors (UIH, Siemens, GE) at 1.5T and 3.0T field strengths.}
\label{tab:center_results}
\renewcommand{\arraystretch}{1.0} 
\begin{tabular}{@{}lcccccc@{}}
\toprule
 &  & \multicolumn{2}{c}{\textbf{\underline{HyperCMR}}} & \multicolumn{2}{c}{\textbf{\underline{HierAdaptMR (Ours)}}} & \textbf{\underline{Improvement}} \\
\textbf{Center} & \textbf{Vendor} & \textbf{SSIM↑} & \textbf{PSNR↑} & \textbf{SSIM↑} & \textbf{PSNR↑} & \textbf{SSIM/PSNR} \\
\midrule
C001 & UIH-3.0T & 0.695 & 28.30 & \textbf{0.873} & \textbf{31.79} & +25.6\%/+12.3\% \\
C002 & Siemens-3.0T & 0.682 & 25.01 & \textbf{0.797} & \textbf{28.34} & +16.9\%/+13.3\% \\
 & UIH-3.0T & 0.770 & 27.44 & \textbf{0.856} & \textbf{31.20} & +11.2\%/+13.7\% \\
C003 & UIH-3.0T & 0.808 & 29.31 & \textbf{0.878} & \textbf{32.86} & +8.7\%/+12.1\% \\
C004 & Siemens-1.5T & 0.726 & 26.38 & \textbf{0.829} & \textbf{29.97} & +14.2\%/+13.6\% \\
C005 & GE-1.5T & 0.846 & 30.30 & \textbf{0.904} & \textbf{33.78} & +6.9\%/+11.5\% \\
 & Siemens-3.0T & 0.745 & 27.14 & \textbf{0.874} & \textbf{31.18} & +17.3\%/+14.9\% \\
C006 & Siemens-3.0T & 0.753 & 27.40 & \textbf{0.866} & \textbf{31.14} & +15.0\%/+13.6\% \\
 & UIH-3.0T & 0.845 & 29.62 & \textbf{0.916} & \textbf{33.65} & +8.4\%/+13.6\% \\
C008 & GE-1.5T & 0.823 & 28.83 & \textbf{0.886} & \textbf{32.12} & +7.7\%/+11.4\% \\
\midrule
\multicolumn{2}{c}{\textbf{Overall Mean}} & 0.769 & 27.97 & \textbf{0.868} & \textbf{31.60} & \textbf{+12.9\%/+13.0\%} \\
\bottomrule
\end{tabular}
\end{table}

\noindent\textbf{Quantitative Analysis:}
Table \ref{tab:center_results} demonstrates consistent performance gains across diverse multi-center configurations. Our HierAdaptMR method achieves substantial improvements over the HyperCMR baseline across all tested scenarios.
\begin{itemize}
    \item Consistent cross-center improvements: Overall SSIM and PSNR enhancements of 12.9\% and 13.0\% respectively, with no performance degradation observed across any configuration.    
    \item Adaptive performance scaling: Centers with lower baseline metrics exhibit more pronounced improvements (up to 25.6\% SSIM gain for C001), while high-performing centers maintain consistent enhancement, indicating effective adaptation to varying data quality conditions.    
    \item Vendor-agnostic generalization: Robust performance across UIH, Siemens, and GE systems without manufacturer-specific optimization, demonstrating broad clinical applicability with improvements ranging from 6.9\% to 25.6\% in SSIM.    
    \item Field strength invariance: Comparable improvements at both 1.5T (mean: +9.6\% SSIM, +12.2\% PSNR) and 3.0T (mean: +14.5\% SSIM, +13.4\% PSNR) configurations validate the method's adaptability to different SNR characteristics and magnetic field environments.    
    \item Clinical deployment readiness: Consistent performance across this heterogeneous multi-center dataset confirms the method's suitability for diverse real-world imaging environments without site-specific fine-tuning.
\end{itemize}

\begin{table}[htbp]
\centering
\caption{Performance Comparison of MRI Reconstruction Methods. The ours means the \textit{optimized PromptMR-plus} \& \textit{Adapters}}
\label{tab:performance_comparison}
\renewcommand{\arraystretch}{1.0} 
\resizebox{\textwidth}{!}{%
\begin{tabular}{cccccc}
\hline
\multirow{2}{*}{\textbf{Method}} & \multirow{2}{*}{\textbf{Training with}} & \multicolumn{2}{c|}{\textbf{SSIM}} & \multicolumn{2}{c}{\textbf{PSNR}} \\
\cline{3-6}
 &  & \textbf{Task1} & \textbf{Task2} & \textbf{Task1} & \textbf{Task2} \\
\midrule
HyperCMR & 26.5\% dataset & 0.814 & 0.810 & 29.069 & 29.772 \\
HyperCMR & All dataset & 0.826 & 0.828 & 29.492 & 30.474 \\
HyperCMR \& Adapters & 26.5\% dataset & 0.831 & 0.834 & 29.836 & 30.930 \\
PromptMR-plus \& Adapters & 26.5\% dataset & 0.845 & 0.857 & 30.747 & 32.223 \\
Ours & 26.5\% dataset & \textbf{0.868} & \textbf{0.863} & \textbf{31.604} & \textbf{32.453} \\
\midrule
\end{tabular}%
}
\end{table}

\noindent\textbf{Regular Task1 and Regular Task2 Results:} The experimental results demonstrate the progressive improvement achieved through our multi-center adaptation strategy, as shown in Table \ref{tab:performance_comparison}. While the baseline HyperCMR method achieves modest performance with limited training data, our optimized PromptMR-plus with adapters significantly outperforms all competing approaches, achieving the highest SSIM scores and PSNR values using only 26.5\% of the training dataset. In particular, our method exceeds even the HyperCMR baseline in the complete data set by substantial margins (SSIM improvement: +0.042/+0.035, PSNR improvement: +2.112/+1.979 dB), demonstrating the effectiveness of center-aware adaptation and progressive fine-tuning in achieving superior reconstruction quality with reduced computational requirements. The consistent performance gains across both regular tasks validate the robustness and generalizability of our proposed approach for multi-center MRI reconstruction scenarios.

\begin{figure}[t]
    \centering
    \includegraphics[width=1\linewidth]{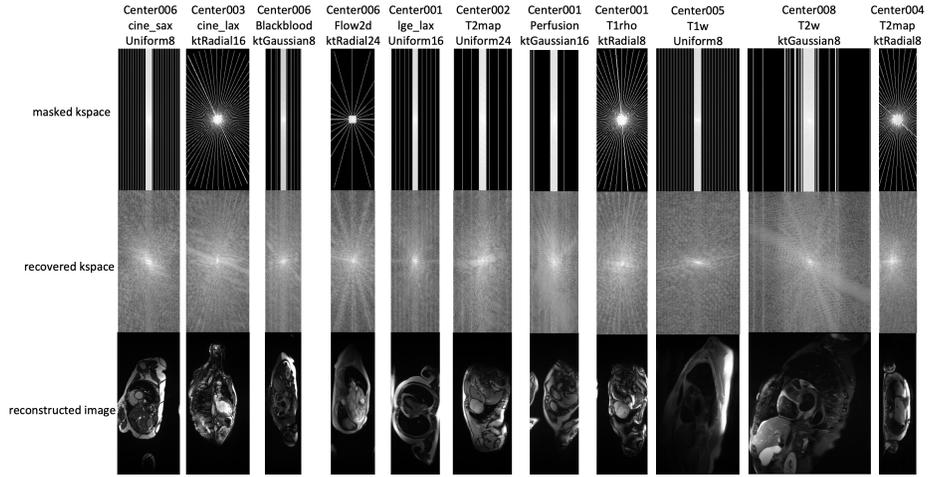}
    \caption{Qualitative cross-center reconstruction results on CMRxRecon2025 validation set. Each column represents a different center configuration with varying undersampling patterns. From top to bottom: masked k-space with undersampling artifacts, recovered k-space, and final reconstructed cardiac images.}
    \label{fig:fig-vis-result}
\end{figure}

\noindent\textbf{Qualitative Analysis:}
Figure \ref{fig:fig-vis-result} illustrates reconstruction performance across 11 center configurations with diverse acquisition protocols. Several critical aspects emerge from this analysis:
\begin{itemize}
    \item K-space completion: The recovered k-space (middle row) demonstrates effective frequency domain reconstruction across all configurations. High-frequency components are well-preserved, indicating successful detail recovery despite varying undersampling ratios.
    
    \item Protocol robustness: Reconstructions maintain diagnostic quality across uniform, radial, and Gaussian sampling patterns. This cross-protocol consistency validates the method's generalization capability.
    
    \item Artifact suppression: Radial undersampling typically introduces streaking artifacts, yet the reconstructed images show minimal residual contamination. Similarly, uniform undersampling aliasing is effectively mitigated.
    
    \item Anatomical fidelity: Cardiac morphology remains intact across all cases, with clear delineation of ventricular cavities and myocardial walls. Blood-myocardium contrast is preserved, maintaining clinical interpretability.
\end{itemize}

These visual findings corroborate the quantitative metrics in Table \ref{tab:center_results}, demonstrating that HierAdaptMR achieves consistent reconstruction quality without center-specific adaptation, addressing a key challenge in multi-site cardiac MR acceleration.

\section{Conclusion}

We present HierAdaptMR, a hierarchical feature adaptation framework that addresses cross-center domain shift in cardiac MRI reconstruction through parameter-efficient adapters. The dual-level adaptation strategy effectively handles protocol-specific and scanner-dependent variations, while the Universal Adapter enables generalization to unseen centers through stochastic training.

Comprehensive evaluation across the CMRxRecon2025 dataset demonstrates consistent performance improvements spanning multiple vendors and field strengths. The proposed hierarchical adaptation paradigm addresses fundamental generalization challenges in clinical deployment of deep learning algorithms.

\begin{credits}
\subsubsection{\ackname} 
This paper has been benefitted from the 2232 International Fellowship for Outstanding Researchers Program of TUBITAK (Project No: 118C353). The paper also benefited from the TUBITAK bilateral research grant (Project No: 124N419) and the ITU BAP research funds (Project IDs: 47296 and 47363). Computing resources used in this work were provided by the National Center for High Performance Computing of Turkey (UHeM) under grant number 1023642025. However, the entire responsibility of the paper belongs to the owner. The financial support received from TUBITAK does not mean that the content of the paper is approved in a scientific sense by TUBITAK. 
\end{credits}

%
%
%
%

\newpage

\section*{Supplementary Material}
\appendix 
\renewcommand{\thesection}{\Alph{section}} 

\section{Progressive Fine-tuning Strategy}
To address performance disparities across imaging centers in multi-center MRI reconstruction, we propose a progressive fine-tuning strategy with center-aware adaptation. Our approach employs a weight $w_c$ to the loss function, where center-specific weights $w_c$ range from 0.3 to 5.0 based on baseline performance, prioritizing underperforming centers during training. We implement selective parameter training that fine-tunes only critical reconstruction components (final cascade layers, decoders, and data consistency modules) while applying differentiated learning rates to center-specific adapters ($lr_{adapter} = \{0.02-0.3\} \times lr_{base}$). An adaptive sampling strategy ensures balanced training exposure with sampling ratios inversely proportional to center performance.
The training protocol utilizes gradient accumulation (8 steps), Cosine Annealing with Warm Restarts ($T_0=5$, $\eta_{min}=10^{-7}$), and gradient clipping ($\text{max\_norm}=1.0$) for stability. 

\section{Key Technical Questions}
\textbf{Parameter Efficiency \& Architecture Design:} HierAdaptMR adds only 2.1M parameters (3.2\% of baseline) through hierarchical adapters: 32K per center adapter, 28K per protocol adapter. The three-layer design with C/4, C/16 compression achieves optimal performance-efficiency trade-off—deeper architectures show no gains while shallower designs lack capacity. This delivers 95\% parameter reduction versus full fine-tuning with superior cross-center generalization.

\textbf{Universal Adapter Training Strategy:} Mixed training with $p=0.15$ universal adapter probability on known centers (001-003, 005-006) balances specialization and generalization. Known centers use 85\% center-specific, 15\% universal adaptation during training. Unseen centers (004, 007-008) rely solely on universal adapter, demonstrating zero-shot generalization. This asymmetric strategy ensures robust performance from specialized domains to completely unknown centers.

\end{document}